\pdfoutput=1
\documentclass[11pt]{article}

\usepackage{naacl2021}

\usepackage{microtype}

\usepackage{adjustbox}
\usepackage{url}
\usepackage{algorithm2e}
\usepackage{amssymb}
\usepackage{amsmath}
\usepackage{arydshln}
\usepackage{verbatim}
\usepackage{booktabs}
\usepackage{graphicx} %package to manage images
\graphicspath{ {./images/} }

\usepackage{booktabs}
\usepackage{xcolor}
\usepackage{multirow}
\definecolor{amber}{rgb}{1.0, 0.49, 0.0}
\definecolor{applegreen}{rgb}{0.55, 0.71, 0.0}

\usepackage{times}
\usepackage{latexsym}
\usepackage{amsfonts}
\usepackage{graphicx} %package to manage images

\usepackage[T1]{fontenc}
\usepackage[utf8]{inputenc}

\usepackage{microtype}

\usepackage{mathtools}

\newcommand{\wv}{{\mathbf w}}
\newcommand{\mv}{{\mathbf m}}
\newcommand{\vv}{{\mathbf v}}
\newcommand{\xv}{{\mathbf x}}
\newcommand{\yv}{{\mathbf y}}
\newcommand{\zv}{{\mathbf z}}
\newcommand{\hv}{{\mathbf h}}

\newcommand{\ours}{LightningDOT }
\newcommand{\oursEOS}{LightningDOT}
\title{LightningDOT: Pre-training Visual-Semantic Embeddings  for \\ Real-Time Image-Text Retrieval}
\author{
    Siqi Sun\thanks{\; Equal Contribution.},\; Yen-Chun Chen\footnotemark[1],\; \\
    {\bf Linjie Li,}\; {\bf Shuohang Wang,}\; {\bf Yuwei Fang,}\; {\bf Jingjing Liu} \\

    Microsoft Corporation\\ % \siqi{Do we change institution name?}\\
    \{siqi.sun, yen-chun.chen, lindsey.li, shuohang.wang, yuwfan, jingjl\}@microsoft.com\\
}

\begin{document}
\maketitle
\begin{abstract}
Multimodal pre-training has propelled great advancement in vision-and-language research. These large-scale pre-trained models, although successful, fatefully suffer from slow inference speed due to enormous computation cost mainly from cross-modal attention in Transformer architecture. When applied to real-life applications, such latency and computation demand severely deter the practical use of pre-trained models. In this paper, we study Image-text retrieval (ITR), the most mature scenario of V+L application, which has been widely studied even prior to the emergence of recent pre-trained models. We propose a simple yet highly effective approach, \ours that accelerates the inference time of ITR by thousands of times, without sacrificing accuracy. \ours removes the time-consuming cross-modal attention by pre-training on three novel learning objectives, extracting feature indexes offline, and employing instant dot-product matching with further re-ranking, which significantly speeds up retrieval process. In fact, \ours achieves new state of the art across multiple ITR benchmarks such as Flickr30k, COCO and Multi30K, outperforming existing pre-trained models that consume 1000$\times$ magnitude of computational hours.\footnote{Code and pre-training checkpoints are available at \url{https://github.com/intersun/LightningDOT}.}
\end{abstract}

\begin{figure}
    \centering
    \includegraphics[width=\linewidth]{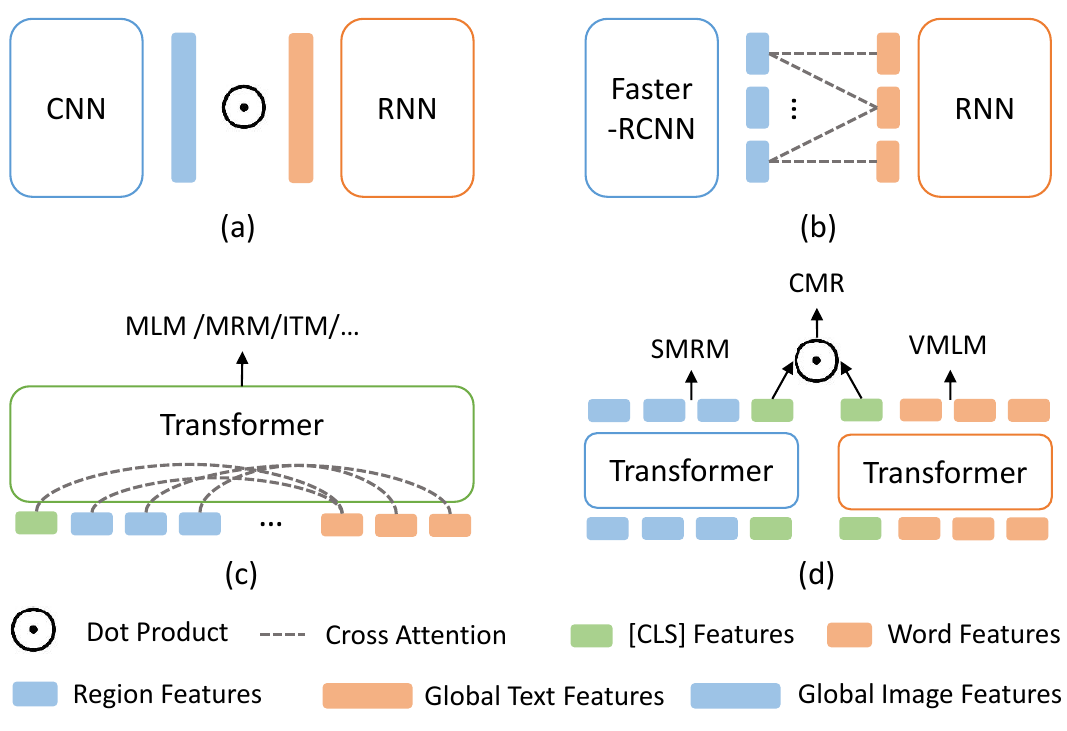}
    \vspace{-6pt}
    \caption{\small{Evolution of Image-Text Retrieval (ITR) paradigm. (a) Early work~\cite{faghri2017vse++} using dot product to learn the similarity between global image features and global text features. (b) Later study~\cite{lee2018stacked} applying cross-attention between the features of each region and each word. (c) Pre-trained V+L models~\cite{chen2020uniter} with deep Transformer. (d) \ours without cross-attention}. CMR, SMRM and VMLM refer to different pre-training tasks, which will be introduced later in method section.}
    \label{fig:intro}
    \vspace{-4mm}
\end{figure}
\section{Introduction}

Image-text retrieval (ITR) has been widely studied as a staple benchmark task in both NLP and computer vision communities. Traditional ITR search engines typically deploy ranking-based models built upon visual-semantic embedding matching~\cite{faghri2017vse++,huang2018learning} or deep cross-modal fusion with attention mechanism~\cite{lee2018stacked,li2019unicoder,li2020oscar}. Earliest works~\cite{kiros2014unifying,faghri2017vse++,wang2018learning} employ separate image encoder (\emph{e.g.,} CNN) and text encoder (\emph{e.g.,} RNN), the embeddings from which are then measured by doc product for similarity matching (Figure~\ref{fig:intro}(a)).
Later studies~\cite{lee2018stacked,lee2019learning,wang2019camp,zhang2020context} improve this paradigm by employing advanced region-level visual encoder (\emph{e.g.,} Faster-RCNN) and applying cross-attention between word features and region features for multimodal fusion (Figure~\ref{fig:intro}(b)).

With the advent of Transformer~\cite{vaswani2017attention} and BERT~\cite{devlin2019bert}, cross-modal retrieval tasks are more recently dominated by vision-and-language (V+L) pre-trained models, such as ViLBERT~\cite{lu2019vilbert}, UNITER~\cite{chen2020uniter}, OSCAR~\cite{li2020oscar}, and VILLA~\cite{gan2020large}.
Large-scale pre-trained models learned from massive corpus of image-text pairs can power heterogeneous downstream tasks that take diverse modalities as inputs (\emph{e.g.,} text, image, video, audio). These models benefit from the self-attention mechanism in Transformer architecture, learning joint image+text embeddings through pre-training objectives such as masked language modeling (MLM) and masked region modeling (MRM) (Figure~\ref{fig:intro}(c)). 

However, the very ingredient that engenders the success of these pre-trained models, \emph{cross-modal attention} between two modalities (through self-attention), also destines the inevitable latency and huge computation cost in training and deploying such massive-scale models. 
For example, UNITER~\cite{chen2020uniter}  builds upon 12/24 Transformer layers, and trains over 10 million image+text pairs. The inference time of such large models with 110 million parameters is 48 seconds on average for text query from COCO dataset~\cite{MSCOCO}, not scalable in real-life applications serving millions of queries per second.

To make real-time ITR possible with low latency, we ask a bold question: can we go back to the beginning, reverting to simple dot product for efficient cross-modal retrieval? To make this retro experiment feasible, we rely on Transformer to pre-train high-quality image and text encoders, but use efficient dot product for multimodal fusion instead of computationally heavy self-attention. To still facilitate effective cross-modal embedding learning, we use a special \texttt{[CLS]} token on both encoders, which transfers the learned embedding from the other modality (Figure~\ref{fig:intro}(d)). We name this new paradigm \emph{\oursEOS}, for its lightening speed benefiting from dot product computation.

\begin{figure*}[t!]
\centering
{\includegraphics[width=\linewidth]{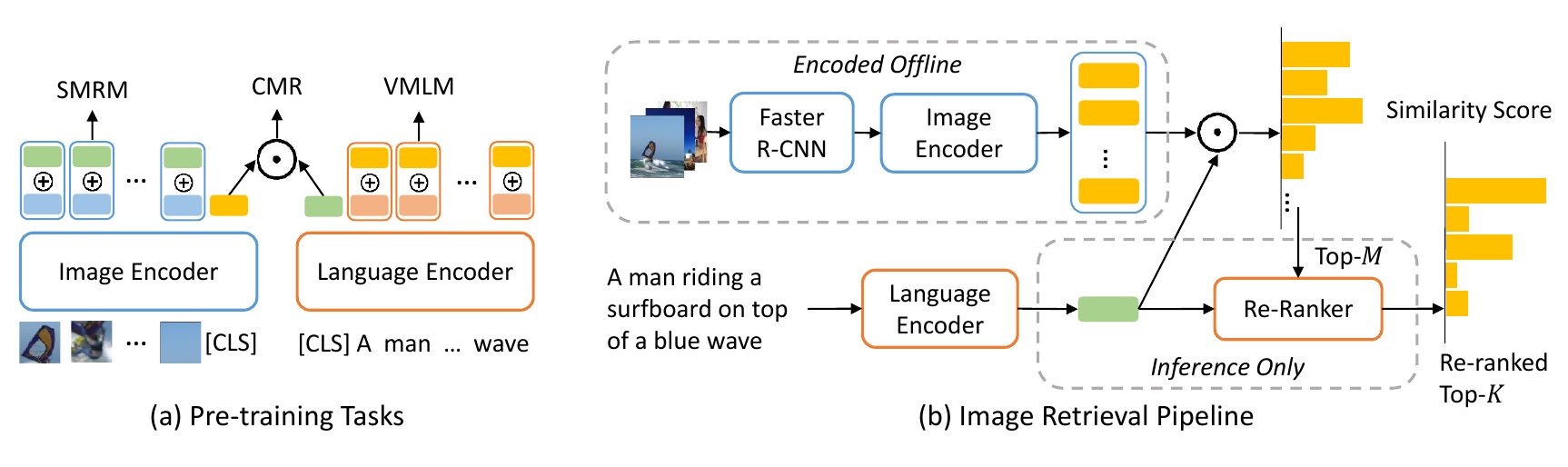}}
\vspace{-6pt}
\caption{\small{An overview of our proposed framework. (a) \ours is pre-trained with Sementic-embedding Fused Mask Region Modeling (SMRM), Visual-embedding Fused Mask Language Modeling (VMLM) and Cross-modal Retrieval (CMR). (b) \ours ITR pipeline (image retrieval as an example). Similarities between input textual query and image candidates are computed via dot product. During inference, image representations can be computed offline, and a re-ranker can be applied for better accuracy, still with significant speedup.}}
\label{fig:model_pipeline}
\vspace{-3mm}
\end{figure*}

By removing the time-consuming cross-attention between modalities, the model can learn visual-semantic embeddings without extensive matching between each image-text pair during inference, as used in existing pre-trained models~\cite{chen2020uniter,li2020oscar,lu2019vilbert}. 
Further, by eliminating the dependency on real-time computation over image-text pairs, we can compute all image and text embeddings independently offline just for once, and reuse these embeddings as cached indexes for new queries on the fly (Figure~\ref{fig:model_pipeline}).

For model training, we propose three learning objectives to jointly train two Transformer blocks: Image Encoder and Language Encoder. Specifically, Visual-embedding fused MLM (namely \emph{VMLM}) and Semantic-embedding fused MRM (namely \emph{SMRM}) ensure cross-modal information is harnessed even without cross-modality self-attention. A cross-modal retrieval objective (namely \emph{CMR}) encourages the model to learn multimodal fusion through pre-training. To maintain competitive model performance, we further introduce a re-ranking mechanism to bring back the benefit of cross-attention methods.

In summary, \ours is designed with late fusion to learn visual-semantic embeddings. Experiments on popular ITR benchmarks show that \ours is 600/1900 times faster than existing pre-trained models on Flickr30k/COCO, while achieving new state-of-the-art results. When retrieving from larger candidate pool (>120K images), \ours is \emph{23,000} times faster. To the best of our knowledge, this is the first known effort on improving V+L model efficiency.

\section{Related Work}

% V+L pre-training
\paragraph{V+L Pre-training} Inspired by the success of Transformer-based~\cite{vaswani2017attention} language model pre-training~\cite{devlin2019bert,liu2019roberta,yang2019xlnet,raffel2020exploring,lan2019albert,clark2020electra}, vision-and-language pre-training~\cite{huang2020pixel, su2019vl,li2020oscar,li2019visualbert} has become the prevailing paradigm in learning multimodal representations, with strong results on tasks such as image-text retrieval~\cite{kiros2014unifying}, visual question answering~\cite{antol2015vqa} and referring expression comprehension~\cite{yu2016modeling}.
Exemplary works include two-stream ~\citep{tan2019lxmert,lu2019vilbert} and single-stream models~\citep{chen2020uniter,li2019unicoder, zhou2019unified}. 
Multi-task learning~\cite{lu201912} and adversarial training~\cite{gan2020large} are also explored.
This family of pre-training methods aims for general-purpose V+L without computation cost consideration. To the best of our knowledge, our work is the first known effort on pre-training visual-semantic embedding that enables low-latency real-time cross-modal retrieval.
Ours is concurrent work with CLIP~\cite{radford2021learning}.

% image retrieval
\paragraph{Image-Text Retrieval} Early cross-modal embedding works~\cite{kiros2014unifying, wang2018learning, faghri2017vse++} focus on using a two-stream model to learn a unified visual-semantic embedding, with progressive improvement on two popular benchmarks: Flickr30K~\cite{plummer2015flickr30k} and COCO~\cite{MSCOCO}.
Later methods with cross-attention~\cite{lee2018stacked, lee2019learning, wang2019camp, zhang2020context} become more popular, with significant performance gain. 
Pre-trained V+L models also fall into this category.
By exploiting large-scale image-text datasets, pre-trained V+L models further push the performance on Flickr30K and COCO. 
Although achieving high recall, cross-attention requires excessive computation cost during inference that cannot be overlooked.\footnote{The total inference time is quadratic to the dataset size with cross-attention for image-text retrieval task.} In this work, inspired by dense retrieval in text retrieval domain~\cite{guu2020realm, karpukhin2020dense, xiong2020approximate, mao2020generation, lewis2020retrieval}, we propose a more efficient attention-less framework. 
With pre-training, our model achieves better performance while being significantly faster than cross-modal attention methods. Note that the proposed approach is orthogonal to model compression techniques that reduce the number of layers/parameters ~\cite{sun2019patient, jiao2019tinybert}, since we do not reduce the number of parameters from the UNITER baseline. These two approaches can be combined to further boost the speed, which is an interesting future work direction.

\section{LightningDOT Framework}

In this section, we present the proposed \ours framework, which consists of two deep Transformers as image and language encoders. We first introduce three tasks designed to pre-train the model, then present our inference pipeline from offline feature extraction to online instant retrieval.

\subsection{Model Pre-training}

We denote the Transformer-based~\cite{vaswani2017attention} image encoder and language encoder by $f_{\theta_V}$ and $f_{\theta_L}$, respectively ($\theta_V$, $\theta_L$ are learnable parameters).
Given a dataset of paired image and text $\{ (i, t) \} $,
we first extract  region features $\vv = \{\vv_0, \vv_1, \dots, \vv_N\}$  ($\vv_j \in \mathbb{R}^{d_v}$, $N$ is the number of regions) for image $i$, along with bounding box positions of regions via a pre-trained Faster-RCNN~\citep{ren2015faster,anderson2018bottom}.\footnote{$\vv_0$ is a special \texttt{[CLS]} embedding.}
The image encoder $f_{\theta_V}$ encodes this sequence of image regions into a $d$-dimensional space $f_{\theta_V}(\vv) = \hv = \{ \hv_0, \dots, \hv_N \}$ $(\hv_j \in \mathbb{R}^{d})$.
The corresponding text $t$ is tokenized into sub-word units and projected into high-dimensional feature vectors $\wv = \{ \wv_0, \wv_1, ..., \wv_T \}$ ($\wv_j \in \mathbb{R}^{d_w}$, $T$ is the number of tokens) following~\citet{devlin2019bert}.\footnote{A 30k BPE~\citep{sennrich2015neural} vocabulary (bert-base-cased) is used to tokenize the text. A special \texttt{[CLS]} token is also prepended following the common practice ($\wv_0$).}
Similarly, the text encoding process can be written as  $f_{\theta_L}(\wv) = \zv = \{ \zv_0, \dots, \zv_T \}$ $(\zv_j \in \mathbb{R}^{d})$. 
We regard the output \texttt{[CLS]} embedding $\hv_0$ as global image representation, and $\zv_0$ as global text representation. Following sections discuss how to jointly train these two encoders to learn strong visual-semantic embeddings, through three pre-training objectives.

\paragraph{Visual-embedding Fused Masked Language Modeling (VMLM)}

Masked Language Modeling (MLM) pre-training is first proposed by~\citet{devlin2019bert}, where 15\% of the words are masked\footnote{In practice, this 15\% is further decomposed into 10\% random words, 10\% unchanged, and 80\% \texttt{[MASK]}.} and the model is trained to reconstruct the masked words. Formally, we denote $\mathbf{w}_\mv = \{ \mathbf{w}_{\mv_{1}}, \dots, \mathbf{w}_{\mv_M} \}$ as masked tokens, where $\mathbf{m}\in \mathbb{N}^M$ is the set of masked indices of size $M$, randomly sampled from a natural number $\mathbb{N}$.  $\mathbf{w}_{\setminus \mathbf{m}}$ are the unmasked words.
MLM can be optimized by minimizing the negative log-likelihood:
\begin{equation}
\label{eq:mlm}
\begin{split}
    \mathcal{L}_{\text{MLM}}(t) = -&\log P_{\theta_L}(\wv_\mv | \wv_{\setminus \mv}) \\
     = - \frac{1}{M} \sum_{k=1}^M &\log P_{\theta_{\text{mlm}}}(\wv_{\mv_k} | \zv_{\mv_k})\,,
\end{split}
\end{equation}
where $\theta_{\text{mlm}}$ is the additional parameters introduced to map hidden states $\zv$ to word probabilities.

Under the V+L setting, the textual input is usually highly correlated with the image. To leverage this cross-modal relation, we propose visual-embedding fused MLM (VMLM), in which the paired image $i$ is considered as additional input when training the model to reconstruct masked tokens in sentence $t$. The loss function of VMLM can be formulated as:
\begin{equation}
\label{eq:imlm}
\begin{split}
    \mathcal{L}_{\text{VMLM}}(t, i) = -\log P_{\theta}(\wv_\mv &| \wv_{\setminus \mv},\, i) \\
     = - \frac{1}{M} \sum_{k=1}^M \log P_{\theta_{\text{mlm}}}(\wv_{\mv_k} &| \zv_{\mv_k} + \hv_0)\,,
\end{split}
\end{equation}
where $\theta = \{\theta_V, \theta_L\}$ and the word probabilities $P_\theta$ are conditioned on the corresponding image $i$ via the global image representation $\hv_0$.
Although VMLM takes a similar mathematical form to the MLM task proposed in UNITER, they differ in two main aspects: 1) \ours uses two separate encoders ($\hv_0$ is computed by $f_{\theta_V}$); and 2) 
visual dependency is explicitly injected to text representations ($\zv_{\mv_k} + \hv_0$), instead of implicitly learned through cross-modal attention.

\paragraph{Semantic-embedding Fused Masked Region Modeling (SMRM)}
Recent works on V+L pre-training~\citep{lu2019vilbert,tan2019lxmert} have shown that \emph{mask-then-reconstruct} pre-training on image regions also helps image+text embedding learning.
Similar to MLM, Masked Region Modeling (MRM) is supervised by:
\begin{equation}
\label{eq:mrm}
\begin{split}
    \mathcal{L}_{\text{MRM}}(i) &= \mathcal{D}_{\theta_{\text{mrm}}}(\vv_\mv,\, f_{\theta_V}(\vv_{\setminus \mv})) \\
     &= \frac{1}{M} \sum_{k=1}^M \mathcal{D}_{\theta_{\text{mrm}}}(\vv_{\mv_k} , \hv_{\mv_k})\, ,
\end{split}
\end{equation}
where $\mathcal{D}$ can be any differentiable distance function.
Among the variants of MRM, we consider Masked Region Feature Regression (MRFR) with L2 distance and Masked Region Classification with KL-Divergence (MRC-kl), due to their proven success in learning V+L representations~\citep{chen2020uniter}.\footnote{In our implementation, no textual inputs are directly concatenated with image regions due to separate encoding of image and text.} 
In MRFR, the $L_2$ distance between two feature vectors $\xv$ and $\yv$ is defined as: 
\begin{equation}
\mathcal{D}_{\theta_{\text{fr}}}(\xv, \yv) = \sum_{k} \| \xv_k - g_{\theta_{\text{fr}}}(\yv_k) \|_2^2\, ,
\nonumber
\end{equation}
where $ \| \cdot \|_2 $ denotes $L_2$-norm, and $g_{\theta_{\text{fr}}}(\cdot)$ is a learnable Multi-layer Perceptron (MLP) with parameters $\theta_{\text{fr}}$.
The KL-divergence $\mathcal{D}_{\text{KL}}$ in MRC-kl measures distance between two probability distributions: 
\begin{equation}
\mathcal{D}_{\theta_{\text{mrc}}}(\xv, \yv) = \sum_k \mathcal{D}_{\text{KL}}( c(\xv_k) \, || \, g_{\theta_{\text{mrc}}}(\yv_k) )\,,
\nonumber
\end{equation}
where $\theta_{\text{mrc}}$ is the parameters of a trainable MLP that maps feature vector $\xv_k$ to the object class distribution  $c(\xv_k)$ predicted by Faster R-CNN.

To incorporate language information encoded in the paired text, we extend MRM to Semantic-embedding fused MRM (SMRM), where the global text representation $\zv_0$ is exploited when reconstructing masked regions. 
\begin{equation}
\label{eq:lmrm}
\begin{split}
    \mathcal{L}_{\text{SMRM}}(i, t) = \mathcal{D}_{\theta_{\text{mrm}}}(\vv_\mv,\, &f_{\theta_V}(\vv_{\setminus \mv}), t) \\
     = \frac{1}{M} \sum_{k=1}^M \mathcal{D}_{\theta_{\text{mrm}}}(\vv_{\mv_k} &, \hv_{\mv_k} + \zv_0)\,.
\end{split}
\end{equation}
The specific variants SMRFR and SMRC-kl can be derived using the corresponding distance function, which is omitted for simplicity.
Note that both the cross-modal fusion introduced in Eqn.~\eqref{eq:imlm} and Eqn.~\eqref{eq:lmrm} uses simple addition without introducing extra parameters from their uni-modal counterpart.
Moreover, the extra parameters $\theta_{\text{mlm}}$ and $\theta_{\text{mrm}}$ is not needed at downstream inference so will not slow down the retrieval.

\paragraph{Cross-modal Retrieval Objective (CMR)}

Beyond image or text focused reconstructive objectives, we also propose a new pre-training task, Cross-modal Retrieval (CMR), to leverage the paired information between image and text. With this learning objective, the model is optimized to promote high similarity score for a matched image-sentence pair $(i, t)$ and vice versa.
The similarity score between query $t$ and image $i$ is defined as:
\begin{align}
\label{eq:sim_score}
S(t, i) = \langle \zv_0, \hv_0 \rangle\, ,
\end{align}
where $\langle \cdot, \cdot \rangle$ denotes the inner product between two vectors, and $\hv_0$ and $\zv_0$ are the output \texttt{[CLS]} embeddings from image encoder $f_{\theta_V}$ and language encoder $f_{\theta_L}$, respectively.

In order to capture both image-retrieval and text-retrieval supervision signals in a single forward-backward pass, we propose a bi-directional variant of contrastive loss. 
Given any matched image-text pair $(i, t)$, we treat text $t$ as the query, sample $n-1$ negative images $\{i_2, i_3, \dots, i_n\}$, and then compute the objective function as: 
\begin{align}
    \mathcal{L}_{\text{IR}}^{(t)} = -\log \frac{e^{S(t, i_1)}} { \sum_{k=1}^{n} e^{S(t, i_k)}}\, ,
    \nonumber
\end{align}
where $t_1 \coloneqq t$.
Similarly, we take image $i$ as query ($i_1 \coloneqq i$), sample $n-1$ negative text, and compute:
\begin{align}
    \mathcal{L}_{\text{TR}}^{(i)} = -\log \frac{e^{S(i, t_1)}} { \sum_{k=1}^{n} e^{S(i, t_k)}}
    \nonumber
\end{align}
to optimize for text retrieval.

\begin{figure}[t!]
\centering
{\includegraphics[width=0.8\linewidth]{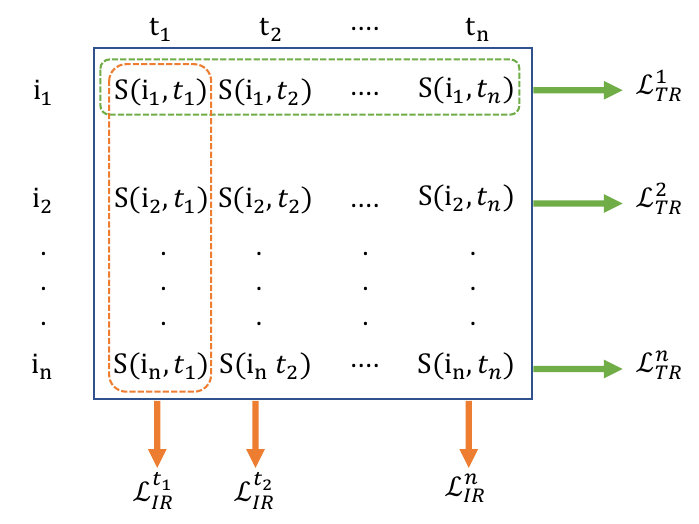}}
\caption{\small{An illustration of the proposed CMR Loss. Note that positive pairs lie in the diagonal of the matrix.}}
\label{fig:in_batch_negatives}
\vspace{-3mm}
\end{figure}

Following \citet{henderson2017efficient, gillick2019learning, karpukhin2020dense}, we use in-batch negatives to avoid the actual sampling of a negative image or text: given a batch of $n$ positive image-text pairs $B=\{(i_1, t_1), \dots, (i_{n}, t_{n})\}$, we use all other images from within the batch as negatives ($\{i_j\}\,, \text{where} \, j \in \{1, 2, \dots, n\} \, \text{and} \, j \neq k$) for every positive pair $(i_k, t_k)$, and vice versa for negative text.
The final CMR loss for batch $B$ is:
\begin{align}
\label{eq:itr}
    \mathcal{L}_{\text{CMR}}(B) = \frac{1}{2n} \sum_{k=1}^{n} \mathcal{L}^{(i_k)}_{\text{TR}} + \mathcal{L}^{(t_k)}_{\text{IR}}\,.
\end{align}
An illustration of $\mathcal{L}_{\text{CMR}}$ is presented in Figure~\ref{fig:in_batch_negatives}.\footnote{The whole similarity matrix can be computed efficiently with one batched matrix multiplication call. This operation can take advantage of GPU hardware with Tensor Cores for faster training.} 
Through joint pre-training with CMR, VMLM and SMRM, the visual-semantic embeddings learned from image encoder and language encoder can be readily applied to downstream tasks. During finetuning stage, we directly adopt CMR loss to supervise the training process.

\subsection{Real-time Inference}

 For simplicity, we take text-to-image retrieval as an example to introduce the real-time inference pipeline (Figure~\ref{fig:model_pipeline}(b)): $(i)$ Offline image feature extraction and encoding; $(ii)$ Online retrieval with text query; and $(iii)$ Online re-ranking with top-retrieved images. Text retrieval is conducted in a symmetric manner.

\paragraph{Offline Feature Extraction}

Image retrieval task requires the model to rank every image $i$ in an image database $I$ based on its similarity to a text query $t$.
In \oursEOS, we first apply the image encoder $f_{\theta_V}$ to all images in $I$, and cache the resulting global image representations $\{\hv_0^{(i)} \in \mathbb{R}^d | i \in I\}$ into an index~\cite{johnson2019billion} in memory for later use. Note that the entire image-to-index process, including Faster-RCNN feature extraction and Transformer encoding, can all be conducted offline. Therefore, for every new query $t$ at real time, the cached index can be reused for maximum inference time saving.

\paragraph{Online Retrieval}

During inference, given a text query $t$, we encode it with the language encoder $\theta_L$, and then compute its similarity score to the embedding of every image in $I$ (stored in memory index) via Eqn~\eqref{eq:sim_score}.
Finally, the images will be ranked by their similarity scores, from the highest to lowest. In practice, people are more interested in top-$K$ retrieval, with a list of $K$ images $I_t$ satisfying:
\begin{multline}
\label{eq:l2_opt}
I_t \coloneqq \{i_{m_k}\}_{k=1}^{K}\,, \, \text{where} \\
S(t, i_{m_1}) \geq  S(t, i_{m_2}) \geq \cdots \geq S(t, i_{m_K}) \hfill \text{and} \\
S(t, i_{m_K}) \geq S(t, i) \hfill \forall i \in (I \setminus I_t)\,.
\end{multline}
This optimization problem has been well studied, and we use FAISS~\citep{johnson2019billion} to solve it in our implementation. It is worth noting that in order to apply fast search, the similarity function has to be \emph{decomposable}. Therefore, we choose the simple dot product as $S$ instead of a more complicated neural network function.
Similarly, for text retrieval, the same architecture can be applied by simply pre-computing the embedding for all sentences and using an image as query instead.

\begin{table*}[t!]
\centering
\small
\setlength{\tabcolsep}{2.7pt}
%\begin{adjustbox}{scale=0.95,center}
%\resizebox{\linewidth}{!}{
\begin{tabular}{lcccccccccccccccc}
\toprule
\multirow{3}{*}{Model} & \multicolumn{7}{c}{COCO Test (5k images)} & \multicolumn{7}{c}{Flickr30K Test (1k images)}  \\ 
\cmidrule(lr){2-8} \cmidrule(lr){9-15}
& \multicolumn{3}{c}{Text Retrieval} & \multicolumn{3}{c}{Image Retrieval} & & \multicolumn{3}{c}{Text Retrieval} & \multicolumn{3}{c}{Image Retrieval} \\
 \cmidrule(lr){2-4} \cmidrule(lr){5-7} \cmidrule(lr){9-11} \cmidrule(lr){12-14}
& R@1 & R@5 & R@10 & R@1 & R@5 & R@10 & AR & R@1 & R@5 & R@10 & R@1 & R@5 & R@10 & AR \\  \midrule

\emph{\textbf{VSE++}}$^*$ & 41.3 &	69.2 &	81.2 &	30.3 &	59.1 &	72.4 &	58.9 &	52.9 &	80.5 &	87.2 &	39.6 &	70.1 &	79.5 &	68.3 \\
\emph{\textbf{SCO}}$^*$ & 42.8 &	72.3 &	83.0 &	33.1 &	62.9 &	75.5 &	61.6 &	55.5 &	82.0 &	89.3 &	41.1 &	70.5 &	81.1 &	69.9\\
\hline
GXN & 42.0	& - &	84.7 & 	31.7 & - &		74.6 & - & 		56.8 & - &		89.6 &	41.5	& - &	80.0 & - \\
SCAN\footnotesize{-single} & 46.4	& 77.4	& 87.2	& 34.4	& 63.7	& 75.7	& 64.1	& 67.9	& 89.0	& 94.4	& 43.9	& 74.2	& 82.8	& 75.4 \\ 
R-SCAN & 45.4 &	77.9 &	87.9 &	36.2 &	65.6 &	76.7 &	65.0 &	66.3	& 90.6	& 96.0	& 51.4	& 77.8	& 84.9	& 77.8
 \\
CAMP & 50.1 &	82.1	& 89.7	& 39.0	& 68.9	& 80.2	& 68.3	& 68.1	& 89.7	& 95.2	& 51.5	& 77.1	& 85.3	& 77.8 \\
CAAN & 52.5	& 83.3 &	90.9 &	41.2 &	70.3 &	82.9 &	70.2 &	70.1 &	91.6 &	97.2 &	52.8 &	79.0 &	87.9 &	79.8\\
\hline
ViLBERT & - & - & - & - & - & - & - & - & - & - & 58.2 & 84.9 & 72.8 & - \\
Unicoder-VL & 62.3 & 87.1 & 92.8 & 46.7 & 76.0 & 85.3 & 75.0 & 86.2 & 86.3 & 99.0 & 71.5 & 90.9 & 94.9 & 88.1 \\
UNITER\footnotesize{-base} & 64.4	& 87.4	& 93.1	& 50.3	& 78.5	& 87.2	& 76.8	& 85.9	& 97.1	& 98.8	& 72.5	& 92.3	& 95.9	& 90.4 \\
UNITER\footnotesize{-large} & 65.7	& 88.6	& 93.8	& 52.9	& 79.9	& 88.0	& 78.1	& 86.9	& 98.1	& \textbf{99.2}	& 75.5	& \textbf{94.0}	& \textbf{96.6} &	91.7 \\
OSCAR & 73.5 & 92.2 & \textbf{96.0} & \textbf{57.5} & \textbf{82.8} & 89.8 & 82.0 & - & - & - & - & - & - & - \\
\hline
\emph{\textbf{\oursEOS}}$^*$ & 60.1	& 85.1 &	91.8 &	45.8 &	74.6	& 83.8	& 73.5	& 83.9	& 97.2	& 98.6	& 69.9	& 91.1	& 95.2	& 89.3 \\ 
+UNITER\footnotesize{base} Re-Ranker & 64.6	& 87.6	& 93.5	& 50.3	& 78.7	& 87.5	& 77.0	& 86.5	& 97.5	& 98.9	& 72.6	& 93.1	& 96.1 &	90.8 \\
+UNITER\footnotesize{large} Re-Ranker & 65.7	& 89.0	& 93.7	& 53.0	& 80.1	& 88.0	& 78.2	& \textbf{87.2}	& \textbf{98.3}	& 99.0	 & \textbf{75.6}	& \textbf{94.0} &	96.5	 & \textbf{91.8} \\
+OSCAR Re-Ranker & \textbf{74.2} &	\textbf{92.4} &	\textbf{96.0} &	57.4 &	82.7 &	\textbf{89.9} &	\textbf{82.1} & - & - & - & - & - & - & -\\
%CAAN 
\midrule
\end{tabular}
%}
%\end{adjustbox}
\caption{
\label{tab:retrival_traditional}
\small{Evaluation results on image-to-text and text-to-image retrieval over Flickr30k and COCO test sets. We compare the proposed method with both task-specific models: VSE++~\cite{faghri2017vse++}, GXN~\cite{gu2018look}, SCO~\cite{huang2018learning}, SCAN~\cite{lee2018stacked}, R-SCAN~\cite{lee2019learning}, CAMP~\cite{wang2019camp} and CAAN~\cite{zhang2020context}, and V+L pre-trained models: ViLBERT~\cite{lu2019vilbert}, Unicoder-VL~\cite{li2019unicoder}, UNITER~\cite{chen2020uniter} and OSCAR~\cite{li2020oscar}. Models in \emph{\textbf{bold}}$^*$ are embedding-based methods without cross-attention.
}}
%\vspace{-3mm}
\end{table*}
\paragraph{Re-ranking}
To further improve retrieval accuracy, we propose  a two-stage approach by adopting an optional re-ranking model. In the first stage, we use \ours to retrieve top-$M$ images (or texts), where $M$ is an integer much smaller than the database (index) size. Next, we apply a stronger retrieval model (usually slower due to the use of cross-attention) to re-rank the retrieved top-$M$ pairs from the first stage.
The final $M$ similarity scores obtained from the second stage will be used to re-compute the desired top-$K$ retrieval ($K \leq M$) in Eqn.~\eqref{eq:l2_opt}. Please refer to figure \ref{fig:model_pipeline} for a more detailed visualization.
Our experiments show that this two-stage approach can benefit from the best of both worlds: maintaining a constant fast speed per query\footnote{The computation time of \ours is negligible compared to that of UNITER. Therefore, the empirical speed is proportional to the number of pairs UNITER has to rank: constant $M$ for \ours $+$ UNITER vs. the whole database (index) size for UNITER only.} while achieving state-of-the-art accuracy. Another advantage of this pipeline is that it can readily incorporate any advanced model as the re-ranker, thus future stronger image-text retrieval models can take advantage of \ours for better efficiency.

\section{Experiments}
This section discusses our experiments on pre-training and evaluating \ours on downstream ITR benchmarks.
\subsection{Datasets and Metrics}

For pre-training, we use pre-processed data provided by~\citet{chen2020uniter}, including 4.2 million images with 9.5 million associated captions from COCO~\citep{MSCOCO}, VG~\citep{krishna2017visual}, Conceptual Captions~\citep{sharma2018conceptual}, and SBU captions~\citep{ordonez2011im2text}.

For evaluation, we use Flickr30k~\cite{plummer2015flickr30k} and COCO~\cite{lin2014microsoft} datasets, which include 31K/123K images, respectively, each associated with 5 human-written captions. Following~\citep{faghri2017vse++}, we split COCO into 114K/5K/5K and Flickr30K into 29K/1k/1k images for train, validation and test.

Downstream performance is measured by recall at $K$ (R@K) for both image and text retrieval tasks.
We also use an additional metric ``AR'', the average of R@K for all $K$ across both image and sentence retrieval tasks.

\begin{table*}[t!]
\centering
%\begin{adjustbox}{scale=0.95,center}
\small
\setlength{\tabcolsep}{2.7pt}
%\resizebox{\linewidth}{!}{
\begin{tabular}{lcccccccccccccccc}
\toprule
\multirow{3}{*}{Model} & \multicolumn{7}{c}{COCO Full (123K Images)} & \multicolumn{7}{c}{Flickr30K Full (31K Images)}  \\ 
\cmidrule(lr){2-8} \cmidrule(lr){9-15}
& \multicolumn{3}{c}{Text Retrieval} & \multicolumn{3}{c}{Image Retrieval} & & \multicolumn{3}{c}{Text Retrieval} & \multicolumn{3}{c}{Image Retrieval} \\
 \cmidrule(lr){2-4} \cmidrule(lr){5-7} \cmidrule(lr){9-11} \cmidrule(lr){12-14}
& R@5 & R@10 & R@20 & R@5 & R@10 & R@20 & AR & R@5 & R@10 & R@20 & R@5 & R@10 & R@20 & AR \\  \midrule
\oursEOS &  40.1 & 51.0 & 62.0 & 28.2 & 37.4 & 47.8 & 44.4 & 69.6 & 78.9 & 86.1 & 51.8 & 62.3 & 72.3 & 70.2 \\ 
+ Re-Ranker\footnotesize{-base} & 47.9 & 58.5 & 67.8 & 35.7 & 45.2 & 55.2 & 51.7 & 74.2 & 81.7 & 88.2 & 56.9 & 66.7 & 75.6 & 73.9  \\
+ Re-Ranker\footnotesize{-large} & \textbf{48.0} & \textbf{59.0} & \textbf{68.9} & \textbf{37.3} & \textbf{46.8} & \textbf{56.4} & \textbf{52.7} & \textbf{75.1} & \textbf{83.9} & \textbf{90.5} & \textbf{60.1} & \textbf{69.5} & \textbf{78.3} & \textbf{76.2} \\
\midrule
\end{tabular}
%}
%\end{adjustbox}
\caption{\label{tab:retrival_full}
\small{Results on the extreme retrieval setting of full Flickr30k and full COCO datasets.}}
%\vspace{-4mm}
\end{table*}
\subsection{Results on Flickr30K and COCO}
We compare the proposed approach with state-of-the-art methods (with and without pre-training) and report the results in Table~\ref{tab:retrival_traditional}. Without cross-attention, our method outperforms non-pre-training approaches by large margins on all metrics. Specifically, our model improves over CAAN~\cite{zhang2020context} (SOTA method with cross-attention) by 3.3\%  (73.5 vs. 70.2) on COCO and 9.5\% (89.3 vs. 79.8) on Flickr30K in terms of AR.  When compared with methods without cross-attention (VSE++~\cite{faghri2017vse++} and SCO~\cite{huang2018learning}), \ours achieves nearly 20-point gain on AR.
Although \ours achieves slightly lower AR than UNITER (pre-training method with cross-attention), with 3.5/1.1 points drop on Flickr30K/COCO, it is 600/1900 $\times$ faster than UNITER during inference time.

We further apply second-stage re-ranking, and use UNITER to score top-$M$ retrieved image-text pairs from \ours to obtain the final top-$K$ ranked lists. With re-ranking, \ours achieves an instant performance lift, surpassing UNITER on both benchmarks, while still 46-95 times faster than UNITER. With an even stronger re-ranker OSCAR, \ours achieves similar results to the state-of-the-art performance on COCO. 
% \begin{table*}[!]
% \centering
% \small
% % \resizebox{\linewidth}{!}{
% \begin{tabular}{lccccc}
% \hline
% Method  & \#images   & UNITER\footnotesize{-base}  & SCAN   & \oursEOS & \oursEOS+re-ranker \\ 
% \hline
% %Flickr30K-test  & 1000   & 0.41   (1$\times$)  & 0.23 (1.8$\times$)  &  0.00064 (639$\times$)  & 0.0089 (46 $\times$)\\
% %COCO-test       & 5000   & 1.95   (1$\times$)  & 1.04 (1.9$\times$)  &  0.00101 (1927$\times$) & 0.020 (95 $\times$)\\
% %Flickr30K-full  & 31014  & 12.8*  (1$\times$)  & 7.10* (1.8$\times$)  &  0.00193 (6591$\times$) & 0.010 (1255 $\times$)\\
% %COCO-full       & 123287 & 48.0*  (1$\times$)  & 25.7* (1.9$\times$)  &  0.00201 (23869$\times$)& 0.021 (2235 $\times$)\\
% Flickr30K-test  & 1000   & 0.41   & 0.23 &  0.00064  & 0.0089 \\
% COCO-test       & 5000   & 1.95     & 1.04   &  0.00101  & 0.020 \\
% Flickr30K-full  & 31014  & 12.8*    & 7.10*   &  0.00193  & 0.010 \\
% COCO-full       & 123287 & 48.0*  & 25.7* &  0.00201 & 0.021 \\
% \hline
% \end{tabular}
% % }
% \caption{\small{Image Retrieval time cost measured by computation time (in seconds) for each query. The computation time for UNITER and SCAN is roughly linear to \#images.}}
% \label{tab:time_cost}
% \end{table*}

\begin{table}[!]
\centering
\small
\setlength{\tabcolsep}{2.7pt}
\begin{tabular}{lcccc}
\hline
Method  & \#images     & SCAN   & Ours & +Re-ranker \\ 
\hline
Flickr30K-test  & 1,000     & 1.8$\times$ &  639$\times$ & 46$\times$ \\
COCO-test       & 5,000      & 1.9$\times$  &  1,927$\times$  & 95$\times$ \\
Flickr30K-full  & 31,014    & 1.8$\times$  &  6,591$\times$  & 1,255$\times$\\
COCO-full       & 123,287  & 1.9$\times$ &  23,869$\times$ & 2,235$\times$\\
\hline
\end{tabular}

\caption{\small{Speedup w.r.t. UNITER-base. We compare \ours (Ours) and +Re-Ranker, plus a lightweight cross-attention method SCAN~\cite{lee2018stacked}. \ours with/without UNITER-base re-ranker is significantly faster.}}
\label{tab:time_cost}
%\vspace{-3mm}
\end{table}

\subsection{Speed \& Space Improvement}
\label{sec:speed}
To demonstrate the efficiency of \oursEOS, we use UNITER-base as baseline to compare inference speed.  We also compare with a more lightweight cross-attention method SCAN~\cite{lee2018stacked}, which uses GRU~\cite{chung2014empirical} instead of a 12-layer Transformer. All methods are tested on a single TITAN RTX GPU, with batch size of 400. As shown in Table~\ref{tab:time_cost}, SCAN is $\sim$1.9$\times$ faster than UNITER-base across both benchmarks, as the computational cost of GRU is much cheaper than that of Transformer (performance drop is significant though). However, the speedup from SCAN is limited, as it computes cross-attention between each query and \emph{all} images. On the other hand,
\ours is 639$\times$ faster than UNITER on Flickr30K. When tested with 5 times more images in COCO, the speedup from \ours is 1927$\times$. Even with re-ranking, \ours is still much more efficient than UNITER-base (46$\times$ faster on Flickr30K and 95$\times$ faster on COCO).

To mimic a real-life scenario for image retrieval, where the candidate pool contains hundreds of thousands of images, we combine all images from training, validation and test set to form a larger candidate pool. Note that models are still trained on the training set.
Although the number of text queries remain the same, the number of candidate images scales up by >20$\times$, where cross-attention methods immediately become impractical. We refer this setting on both benchmarks as Flickr30k-full (31k) and COCO-full (123k). Our algorithm is 6,591$\times$ faster on Flickr30k-full and 23,869$\times$ faster on COCO-full, which clearly shows the advantage of \ours and its potential in real-world applications. With re-ranking, \ours is still more than 1,000$\times$ and 2,000$\times$ faster on Flickr30k-full and COCO-full, respectively. In general, for other re-rankers such as OSCAR, our algorithm can approximately speed up inference by $N_{\text{images}}/M$ times, where $N_{\text{images}}$ is the number of candidate images, and $M$ is number of re-ranked images from top-$M$ retrieved results by \oursEOS.

Similarly, we construct a full setting for text retrieval by combining all text queries from training, validation and test set. Results are summarized in Table~\ref{tab:retrival_full}. Considering the size of candidate pool has become more than 20$\times$ larger, we adopt recall at top 5, 10, 50 as evaluation metrics. Our method achieves reasonably good performance, with AR of 44.4 on COCO and 70.2 on Flickr30K. Re-ranking further lifts AR to 56.4 and 76.2. Results from UNITER or SCAN are not included as the computation of pairwise scores is extremely expensive, given the excessive amount of retrieval candidates. While \ours only takes minutes to evaluate, UNITER-base is estimated to take about 28 days\footnote{This estimation is based on the inference time taken by UNITER-base on a smaller dataset.} to evaluate under the full setting for both image retrieval and text retrieval. 

In addition, We compare all models with the same setting: cache as much as possible for fastest speed, where our model outperforms others in both speed and space on image retrieval. The proposed algorithm maps each image to a 768-dimensional vector, which only consumes about 300Mb storage space for the whole COCO dataset. For cross-attention models such as SCAN, UNITER or OSCAR, they also need to cache image features, which typically requires to save a 36 x 2048 dimensional vector per image, and it consumes about 28GB storage space for COCO dataset.

\begin{table}[t!]
\centering
%\begin{adjustbox}{scale=0.95,center}
\small
\setlength{\tabcolsep}{1pt}
\begin{tabular}{lccccccccc}
\toprule
& \multicolumn{3}{c}{Text Retrieval} & \multicolumn{3}{c}{Image Retrieval} \\
 \cmidrule(lr){2-4} \cmidrule(lr){5-7} 
Method & \small{R@1} & \small{R@5} & \small{R@10} & \small{R@1} & \small{R@5} & \small{R@10} & \small{AR} \\  \midrule
% + 2-layer RANDOM & -	& -	& -	& 51.5	& 80.5	& 88.5 & - \\
% + 2-layer UNITER & - & - & - & 53.6 & 82.2 & 88.8 & - \\
% + 12-layer UNITER & 73.4 &	92.5 &	95.6 &	59.5 &	84.5 &	90.3 &	82.6 \\
% PT ZS & 64.7 &	88.3 &	93.8 &	49.8 &	77.7 &	86.1 &	76.7 \\
R-CNN only & 62.2	& 85.9	& 91.1	& 42.0	& 70.9	& 80.3	& 72.1 \\
+Image Encoder & 73.4 &	92.5 &	95.6 &	59.5 &	84.5 &	90.3 &	82.6 \\
+PT$^\dagger$ & 83.5 &	\textbf{96.4} &	\textbf{98.7} &	68.6 &	\textbf{90.5} &	\textbf{94.8} &	88.8 \\
\oursEOS & \textbf{85.2}	& \textbf{96.4}	& \textbf{98.7}	& \textbf{69.9}	& 90.4	& 94.5	& \textbf{89.2} \\
\midrule
\end{tabular}

\caption{
\label{tab:ablate_flickr}
\small{Ablation studies on model design over Flickr30K validation set. PT$^\dagger$ indicates pre-training with MLM+MRM+CMR, while \ours is pre-trained with VMLM+SMRM+CMR.
}}
%\vspace{-2mm}
\end{table}
\begin{table}[t!]
\centering
%\begin{adjustbox}{scale=0.95,center}
\small
\setlength{\tabcolsep}{1.7pt}
\begin{tabular}{lccccccccc}
\toprule
& \multicolumn{3}{c}{Text Retrieval} & \multicolumn{3}{c}{Image Retrieval} \\
 \cmidrule(lr){2-4} \cmidrule(lr){5-7} 
\oursEOS & \small{R@1} & \small{R@5} & \small{R@10} & \small{R@1} & \small{R@5} & \small{R@10} & \small{AR} \\  \midrule
% R-CNN only & 62.2	& 85.9	& 91.1	& 42.0	& 70.9	& 80.3	& 72.1 \\
No PT & 73.4 &	92.5 &	95.6	& 59.5	& 84.5	& 90.3	& 82.6 \\
%VMLM+SMRM+FT \\
PT(CMR) & 75.0 &	93.9 &	\textbf{97.3} &	61.5 &	85.5 &	91.1 &	84.0 \\
PT(All) & \textbf{78.1} &	\textbf{94.0} &	96.9 &	\textbf{62.6} &	\textbf{85.7} &	\textbf{91.8} &	\textbf{84.8} \\
\midrule
\end{tabular}

\caption{
\label{tab:ablation}
\small{Ablation studies on pre-training tasks over Flickr30K validation set after finetuning on the corresponding training set. All pre-training experiments are conducted on COCO dataset only. PT is short for pre-training. PT(CMR) refers to pre-training using CMR task only, and PT(All) refers to pre-training with all of the three tasks.
}}
%\vspace{-3mm}
\end{table}
\subsection{Ablation Studies}
We conduct ablation studies on Flickr30K (Table~\ref{tab:ablate_flickr}) and compare \ours(L4) against 3 ablated instances: ($i$)``R-CNN only'' (L1): image representations are extracted from Faster R-CNN directly, with no image encoder applied; ($ii$) ``+Image Encoder'' (L2): regional features are encoded with a 12-layer Transformer as the image encoder; ($iii$) ``+PT$^\dagger$'' (L3): our model is pre-trained with MLM+MRM+CMR, then finetuned on Flickr30K. Note that the difference between MLM vs. VMLM and MRM vs. SMRM is whether the predictions of masked tokens (regions) rely on infused embeddings from the other modality. 

Results show that ``R-CNN only'' is not sufficient in learning good image representations for ITR task, while image encoder with Transformer architecture can effectively learn contextualized image representations, hence achieving better performance. Pre-trained models (L3-4) generally achieve better performance, compared to non-pretrained models (L1-2). Comparing ``+PT$^\dagger$'' to the full instance of \oursEOS, dependency on the other modality in VMLM and SMRM brings universal performance lift across all metrics. This indicates that these cross-modal dependencies introduced by VMLM and SMRM are effective in learning the association between image and text inputs.

\begin{table}[!]
\centering
\small
\small
\setlength{\tabcolsep}{3.5pt}
\begin{tabular}{lcccccc}
\hline
& \multicolumn{3}{c}{Multi30K} & \multicolumn{2}{c}{COCO} & \\
 \cmidrule(lr){2-4} \cmidrule(lr){5-6}
Method  & DE   & FR  & CS   & ZH & JA & Meta-Ave\\ 
\hline
% EmbN\cite{EMnb}  & 60.3 & 54.8 & 46.3  & 73.2 & 73.5 & 65.3\\
% PAR.EmbN \cite{Par.EMnb}  & 62.6 & 60.6 & 54.1 & 76.0 & 74.8 & 67.9\\
S-LIWE  & 72.1 & 63.4 & 59.4  & 73.6 & 70.0 & 67.7\\
MULE & 64.1 & 62.3 & 57.7  & {\color[HTML]{0000FF}75.9} & {\color[HTML]{0000FF}75.6} & 67.1 \\
SMALR  & 69.8 & 65.9 & 64.8 & {\color[HTML]{0000FF}77.5} & {\color[HTML]{0000FF}76.7} & 70.9\\
\hline
$\text{M}^3\text{P}$  & 82.0 & 73.5 & 70.2  & 81.8 & \textbf{86.8} & 78.9\\
\hline
UNITER & 85.9 & \textbf{87.1} & 85.7 & \textbf{88.4} & 85.9 & 86.6 \\
\oursEOS &  83.3 &  83.7 &  82.2  & 87.2&  82.3  & 83.7 \\ 
+Re-Ranker & \textbf{86.1} & \textbf{87.1} & \textbf{86.2} & \textbf{88.4} & 86.1 & \textbf{86.8} \\
\hline
\end{tabular}

\caption{\small{Evaluation on multilingual image-text retrieval over Multi30K and COCO datasets.  
We compare with task-specific methods: S-LIWE~\citep{S-LIWE}, MULE~\citep{MULE}, SMALR~\citep{SMALR}, pre-trained method M$^3$P~\cite{huang2020m3p} and UNITER with \emph{translate-test}. 
Numbers in {\color[HTML]{0000FF}blue} indicate the use of different dev/test splits of COCO compared to other methods. UNITER and Re-ranker are large model size.}}
\label{tab:multilingual}
%\vspace{-3mm}
\end{table}

In addition, we investigate the effectiveness of each pre-training task in Table~\ref{tab:ablation}. Comparing to baseline without pre-training, 
pre-training with CMR alone lifts $+1.4$ on AR.
Pre-training with all three tasks achieves the best performance, indicating that the learning of contextualized word and region representations promotes better global alignment between image and text, and these three pre-training tasks work collaboratively to yield better visual-semantic embeddings.

\begin{figure*}[h!]
\centering
{\includegraphics[width=0.9\linewidth]{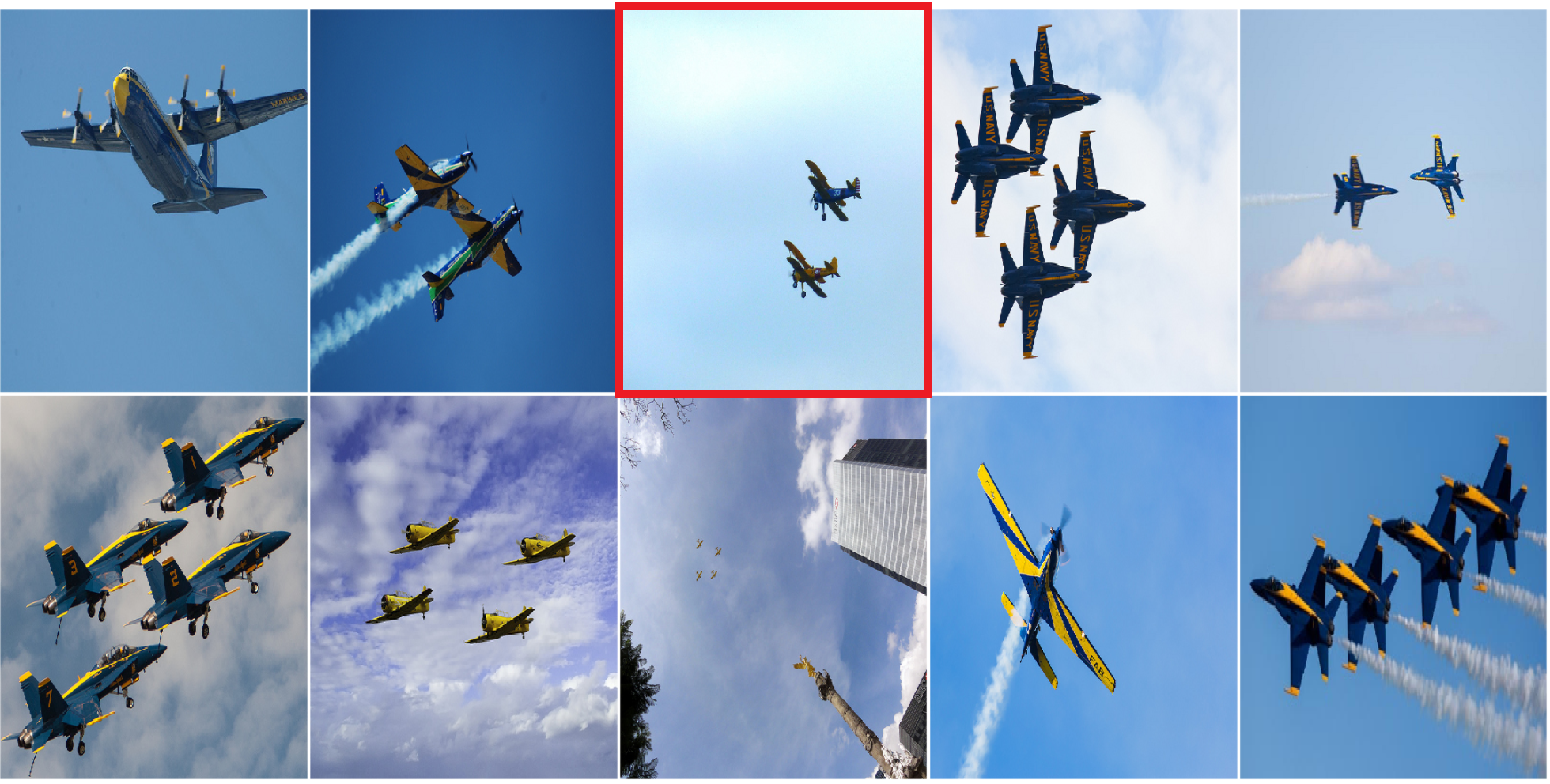}}
\vspace{-6pt}
\caption{Retrieved top 10 images from the query "Sky view of a blue and yellow biplane flying near each other." The ground truth is in the red rectangle.}
\label{fig:example_3_part}
\vspace{-3mm}
\end{figure*}

\subsection{Multilingual Image-Text Retrieval}
We further report results on multilingual image-text retrieval tasks. Specially, we evaluate \ours under the \emph{translate-test} setting, which is to translate the test captions in other languages to English by leveraging Machine Translation (MT) tool.\footnote{We use Microsoft Azure Translation API Service.} Note that our method is only trained on English captions, without exploiting the original or translated captions from multilingual benchmarks.

 We consider two benchmarks: Multi30K~\citep{multi30k, multi30k_extension_1, multi30k_extension_2} with captions in German, French and Czech; and COCO Japanese~\citep{MSCOCO_JA} and Chinese~\citep{MSCOCO_ZH}. 

 Average Recall (AR) is used as the evaluation metric. Meta-Ave, the average of AR over different languages across two benchmarks, is used as a global metric. More details on multilingual ITR benchmarks are included in Appendix.

We compare \ours against 3 task-specific methods: S-LIWE~\citep{S-LIWE}, MULE~\citep{MULE} and SMALR~\citep{SMALR}, which all exploit captions in different languages to learn multilingual or language-agnostic word embeddings. We also compare with a pre-trained model M$^3$P~\citep{huang2020m3p}, which is alternatively pre-trained with image-caption pairs labeled in English and cross-lingual corpus in 100 different languages. Note that all methods discussed above are trained/finetuned on captions in different languages. For fair comparison, we report performance of UNITER under the same translate-test setting, which is finetuned with English captions only and tested on translated captions.

Table~\ref{tab:multilingual} shows similar trends of performance improvements as on English benchmarks. Compared to both state-of-the-art task-specific methods and pre-trained models, \ours under \emph{translate-test} setting achieves new state of the art on most languages and establishes a strong baseline for future study on these multilingual benchmarks.

\subsection{Qualitative Examples}
We show an example of image retrieval results here at figure~\ref{fig:example_3_part} for query as "Sky view of a blue and yellow biplane flying near each other". In addition to the ground truth image in the red rectangle, all the 10 images retrieved by our model are valid retrieval since multiple keywords ("sky", "blue", "yellow", "airplane", "near") are captured for each image. Please see the appendix~\ref{sec:more_ex} for more examples.

\section{Conclusion}
In this paper, we propose a pre-training framework that learns joint visual-semantic embedding without any cross-attention between modalities. \ours outperforms previous state of the art, while significantly speeding up inference time by 600-2000$\times$ on Flickr30K and COCO image-text retrieval benchmarks. Future work includes extending the efficient training framework to other V+L tasks.

\bibliography{anthology,custom}
\bibliographystyle{acl_natbib}

\appendix
\clearpage

\section{Appendix}
\label{sec:appendix}
\subsection{Implementation Details}
To further facilitate the reproductivity of our proposed method, we include more details about the choice of model size and hyper-parameters for both pre-training and fine-tuning.

The model dimensions are set to (L=12, H=768, A=12) for both image encoder and language encoder, where L is the number of stacked Transformer blocks; H stands for hidden activation dimension, and A is the number of attention heads. The total number of parameters in \ours is 220M. Pre-training and finetuning learn the parameters of both encoders. During inference, with offline representation caching, only the forwarding pass with one encoder from the query modality will be performed online.

For both pre-training and finetuning, AdamW~\citep{loshchilov2017decoupled} is used to optimize the model training, with $\beta_1{=}0.9$, $\beta_2{=}0.98$. We adopt a learning rate warmup strategy, where the learning rate is linearly increased during the first 10\% of training steps, followed by a linear decay to 0. We set the L2 weight decay to be 0.01.

During pre-training, we follow UNITER~\cite{chen2020uniter} to randomly sample 1 task per mini-batch update.\footnote{Code obtained from https://github.com/ChenRocks/UNITER.}
Our best model is pre-trained on VMLM+SMRM+CRM for 300,000 optimization steps. We set the batch size to 10240 per GPU (batch size is specified by \#tokens + \#regions, as in UNITER). Pre-training experiments are conducted on 8$\times$ V100 GPUs with 6-step gradient accumulation, and the learning rate is set to be 5e-5. For ablation studies presented in Table 5, the ablated instances of our model are pre-trained for 30k steps on COCO dataset~\cite{lin2014microsoft} only, and the same choice of learning rate and batch size are applied as in the best pre-training setting.

For finetuning, we set batch size $n$ to 96 ($n$ is in examples, instead of the sequence length of tokens and regions), and search learning rate from \{1e-5, 2e-5, 5e-5\}. We select models based on their AR on the validation set. The best learning rate is 5e-5 for COCO  and 1e-5 for Flickr30K. Our models are trained for 15 epochs on Flickr30k, and 20 epochs on COCO. For re-ranking, we choose $k$ from \{20, 50\}.

\subsection{Multilingual Image-Text Retrieval Benchmarks}
When evaluating on ITR under the multilingual setting, we consider two benchmarks: Multi30K~\citep{multi30k, multi30k_extension_1, multi30k_extension_2} and COCO Japanese~\citep{MSCOCO_JA} and Chinese~\citep{MSCOCO_ZH}. 
 Multi30K is constructed by manually translating English captions in Flickr30K~\citep{plummer2015flickr30k} to German, French, and Czech. Each image in Multi30K is paired with 5 captions in German, 1 caption in French and Czech. We adopt the same train/val/test split as in Flickr30K. 
 COCO Japanese~\citep{MSCOCO_JA} collected  820K Japanese captions for 165K COCO images~\citep{lin2014microsoft}. We use the same train/dev/test splits for COCO Japanese as in~\citet{karpathy2015deep}, and present results on the 1K test set. Similarly,~\citet{MSCOCO_ZH} collected 1-2 Chinese captions per image for 20K COCO images to build COCO Chinese.  We follow the original split defined in~\citet{MSCOCO_ZH}. 
 
 \begin{table*}[t!]
\centering
\small
% \resizebox{\linewidth}{!}{
\begin{tabular}{lccccc}
\hline
Method  & \#images   & UNITER\footnotesize{-base}  & SCAN   & \oursEOS & \oursEOS+Re-ranker \\ 
\hline
Flickr30K-test  & 1000   & 0.41     & 0.23   &  0.00064  & 0.0089 \\
COCO-test       & 5000   & 1.95     & 1.04  &  0.00101 & 0.020 \\
Flickr30K-full  & 31014  & 12.8*    & 7.10*   &  0.00193  & 0.010 \\
COCO-full       & 123287 & 48.0*  & 25.7*  &  0.00201 & 0.021 \\
% Flickr30K-test  & 1000   & 0.41   & 0.23 &  0.00064  & 0.0089 \\
% COCO-test       & 5000   & 1.95     & 1.04   &  0.00101  & 0.020 \\
% Flickr30K-full  & 31014  & 12.8*    & 7.10*   &  0.00193  & 0.010 \\
% COCO-full       & 123287 & 48.0*  & 25.7* &  0.00201 & 0.021 \\
\hline
\end{tabular}
% }
\caption{\small{Image retrieval time cost measured by computation time (in seconds) for each query. The computation time for UNITER and SCAN is roughly linear to \#images. Numbers with * are estimated by running time on test set.}}
\label{tab:time_cost_appendix}
\end{table*}

 \subsection{Inference Time}
 We present the detailed inference time of UNITER-base, SCAN the proposed \ours and \ours with UNITER-base re-ranker in Table \ref{tab:time_cost_appendix}, measured by seconds/query. UNITER clearly is the slowest, as the 12-layer Transformer model inference needs to be run between each query and \emph{all} images. Comparing between Flickr30k-test and COCO-test, its inference time scales up linearly with the number of images. 
With the lightweight GRU~\cite{chung2014empirical}, SCAN is $\sim$1.9$\times$ faster than UNITER.
Across all settings, \ours is significantly faster than both cross-attention methods (UNITER-base and SCAN). When adding UNITER-base as the re-ranker, our method slows down by  $\sim$10, but still achieves decent speedup. 

\begin{figure*}[t!]
\centering
{\includegraphics[width=\linewidth]{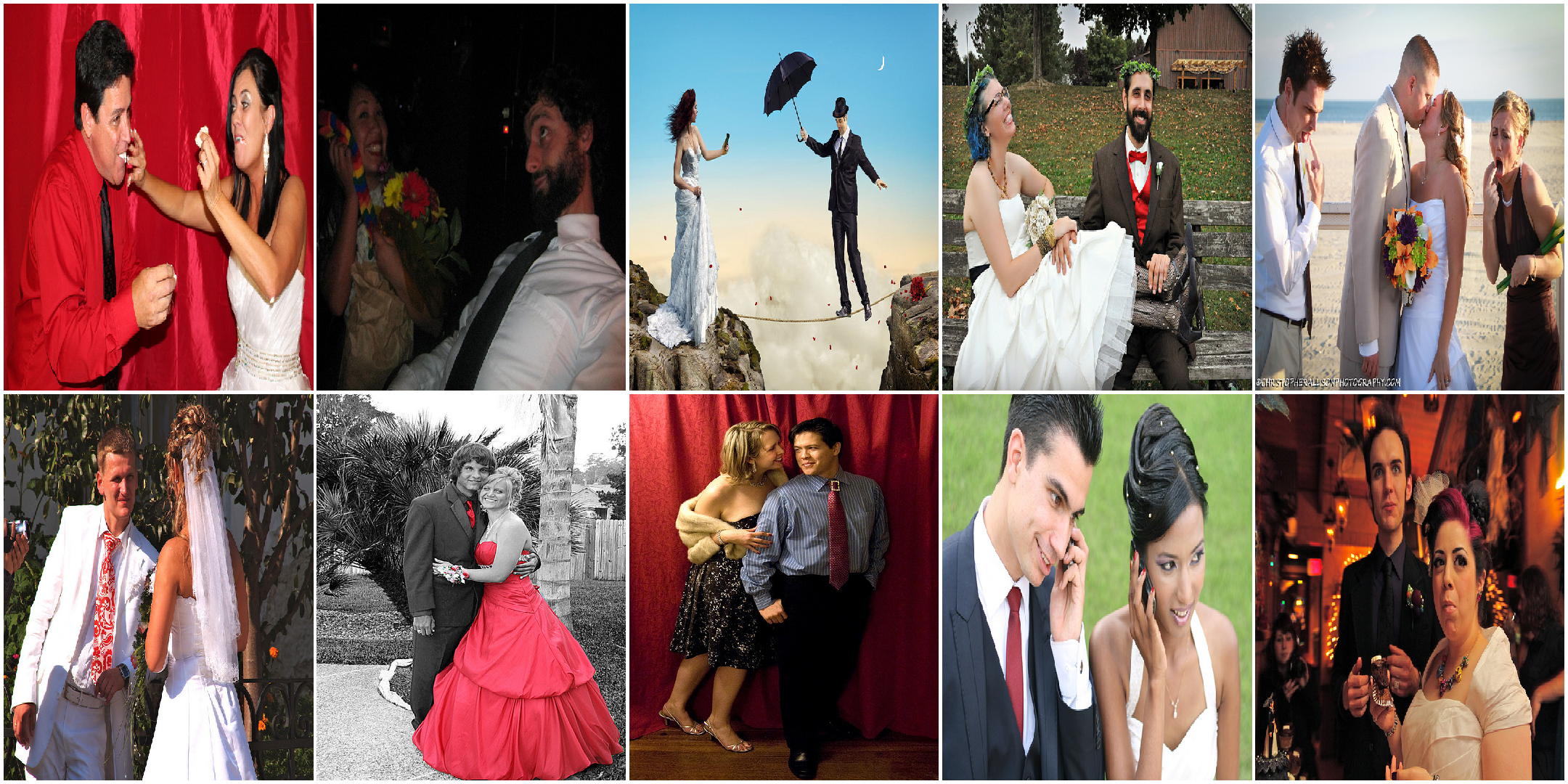}}
\vspace{-6pt}
\caption{Retrieved top-10 images for query "romantic".}
\label{fig:example_1}
\vspace{-3mm}
\end{figure*}

\begin{figure*}[h!]
\centering
{\includegraphics[width=\linewidth]{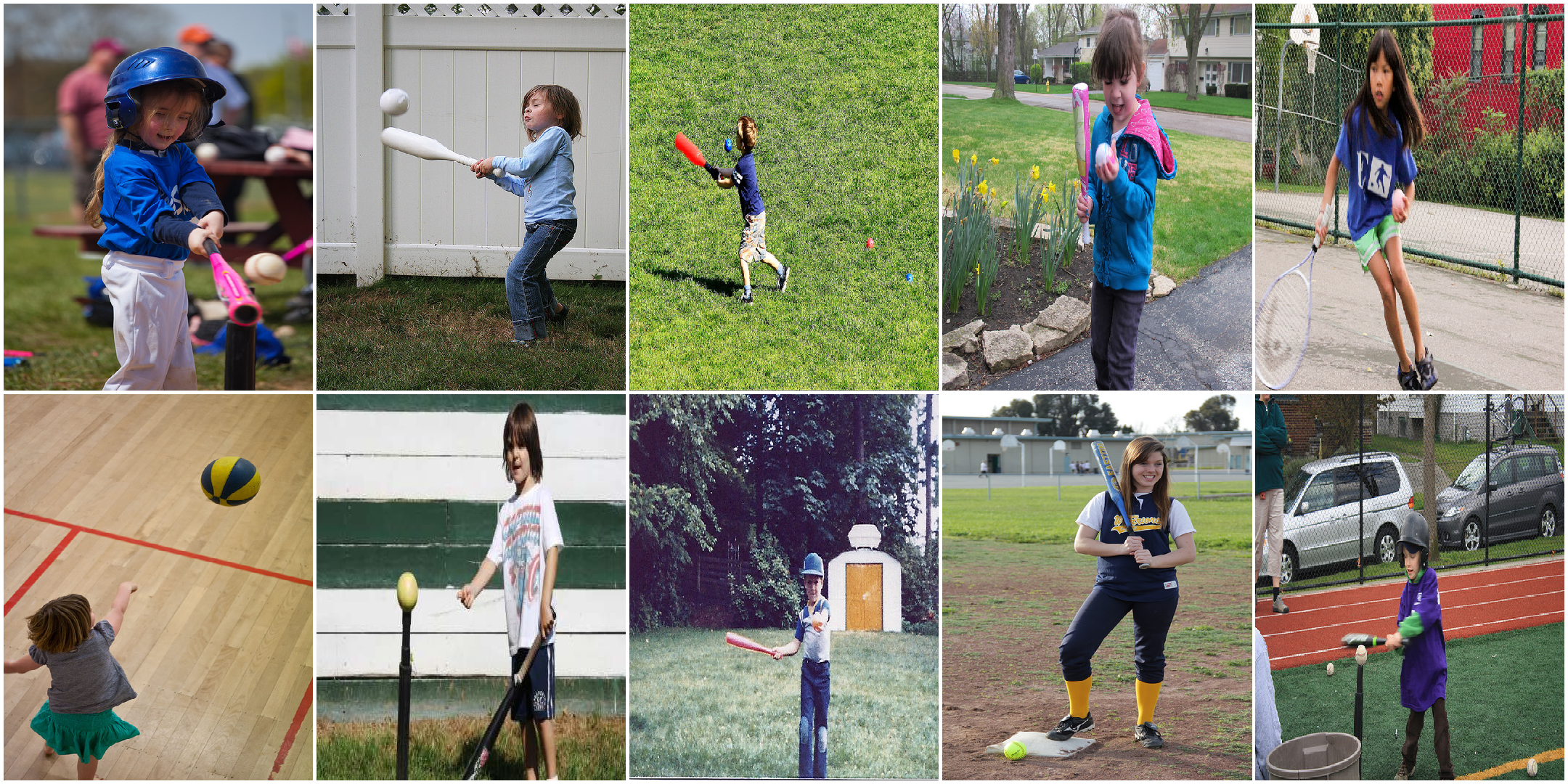}}
\vspace{-6pt}
\caption{Retrieved top-10 images for query "blue girl boy ball"}
\label{fig:example_2}
\vspace{-3mm}
\end{figure*}

\subsection{More Qualitative Examples}
\label{sec:more_ex}
We show several qualitative results of image retrieval (top-10). All results are retrieved from COCO-Full dataset (123k images in total).
Our model can well understand the underlying semantic meaning. For example, ``romantic'' only appears twice in the whole COCO dataset annotations, yet the top retrieved images are all topic-related (Figure~\ref{fig:example_1}).
With multiple keywords, our model attempts to retrieve the combinations of them (if not all). For example, for the query ``blue girl boy ball'' with four keywords, our model retrieves images that capture at least three keywords (Figure~\ref{fig:example_2}).

We also present image retrieval results where the text query is sampled from COCO dataset. We randomly sample 3 queries and present the results as below (ground truth on the top, retrieved top-10 images at the bottom). Clearly, our model retrieves related images from the full dataset.

\begin{figure*}[h!]
\centering
{\includegraphics[width=0.9\linewidth]{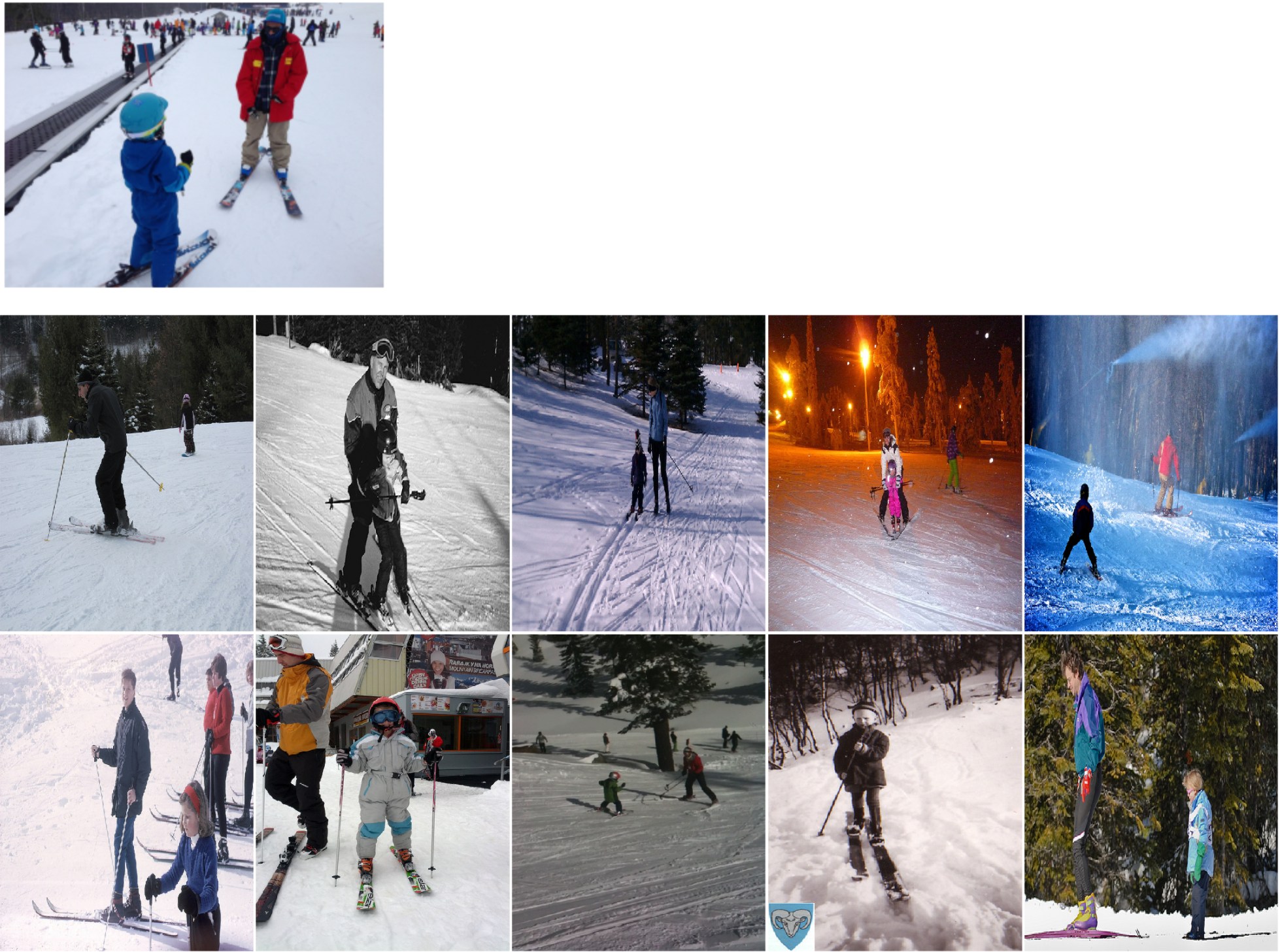}}
\vspace{-6pt}
\caption{Retrieved top 10 images from the query "A man and a little boy on skis on a ski hill." (Top picture is the ground truth.)}
\label{fig:example_3}
\vspace{-3mm}
\end{figure*}

\begin{figure*}[h!]
\centering
{\includegraphics[width=0.9\linewidth]{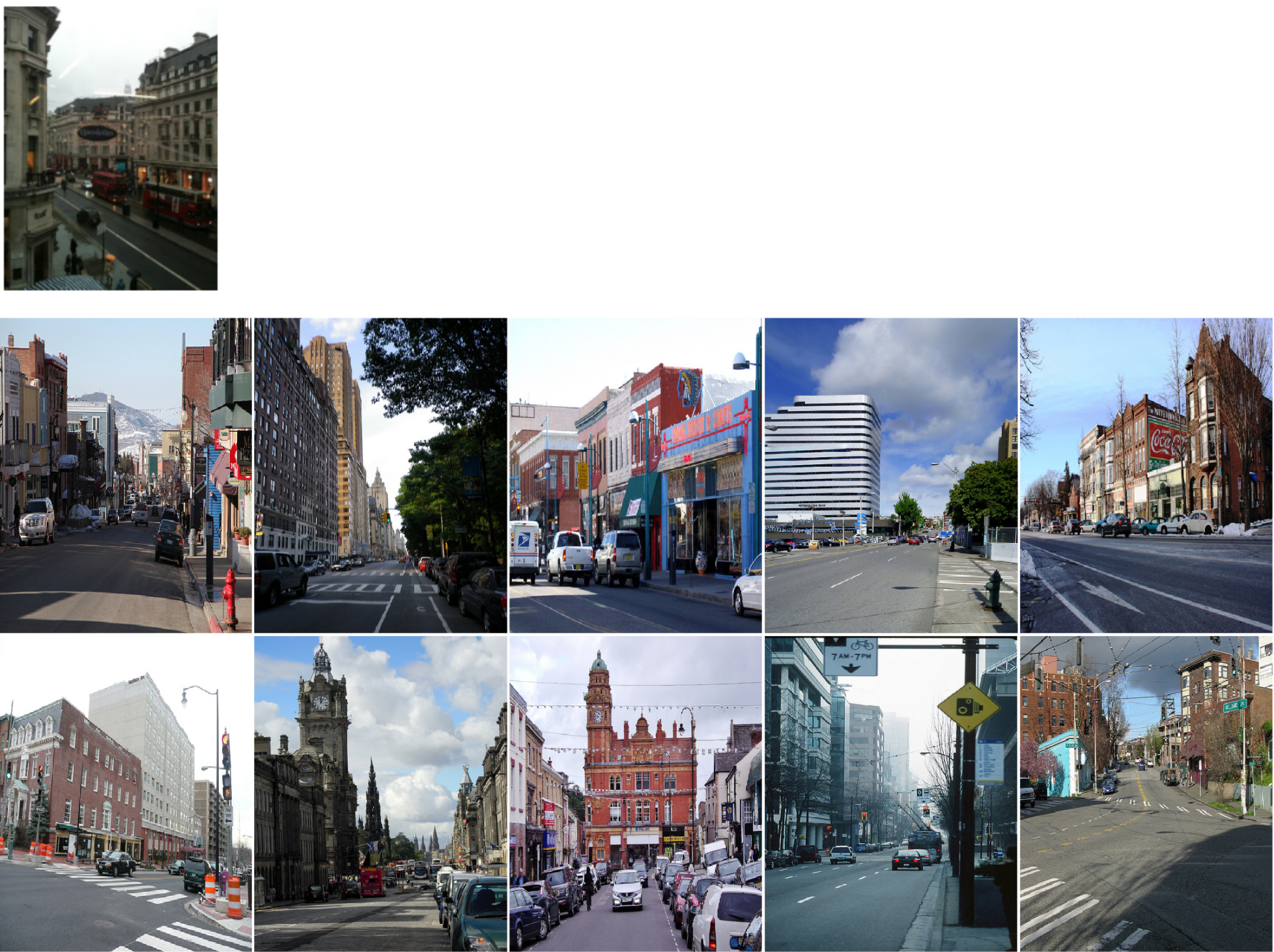}}
\vspace{-6pt}
\caption{Retrieved top 10 images from the query "A road is lined with buildings and has cars on it." (Top picture is the ground truth.)}
\label{fig:example_4}
\vspace{-3mm}
\end{figure*}

\begin{figure*}[h!]
\centering
{\includegraphics[width=0.9\linewidth]{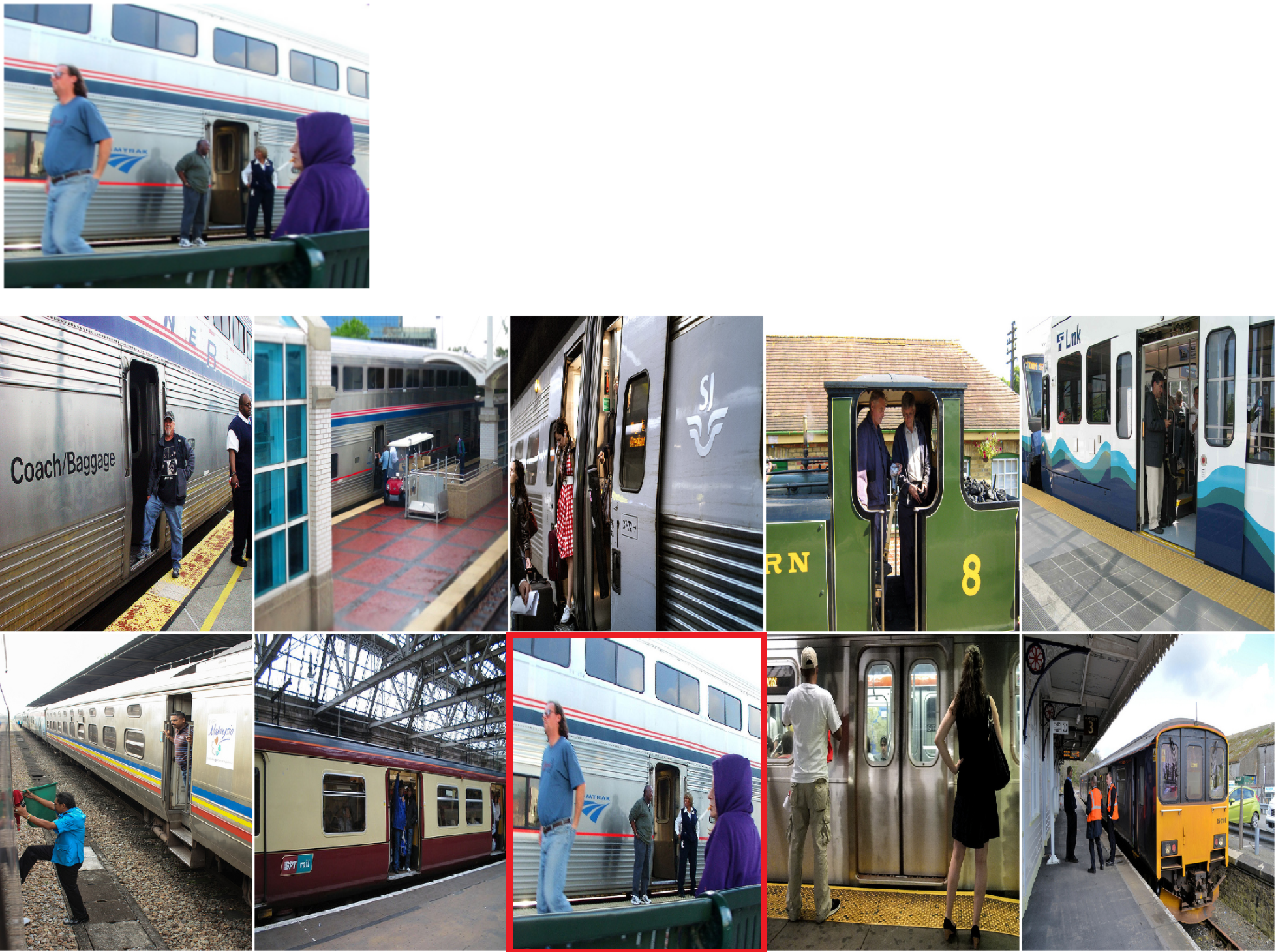}}
\vspace{-6pt}
\caption{Retrieved top 10 images from the query "Two train employees stand near the open train car door." (Top picture is the ground truth.)}
\label{fig:example_5}
\vspace{-3mm}
\end{figure*}

\begin{figure*}[h!]
\centering
{\includegraphics[width=0.9\linewidth]{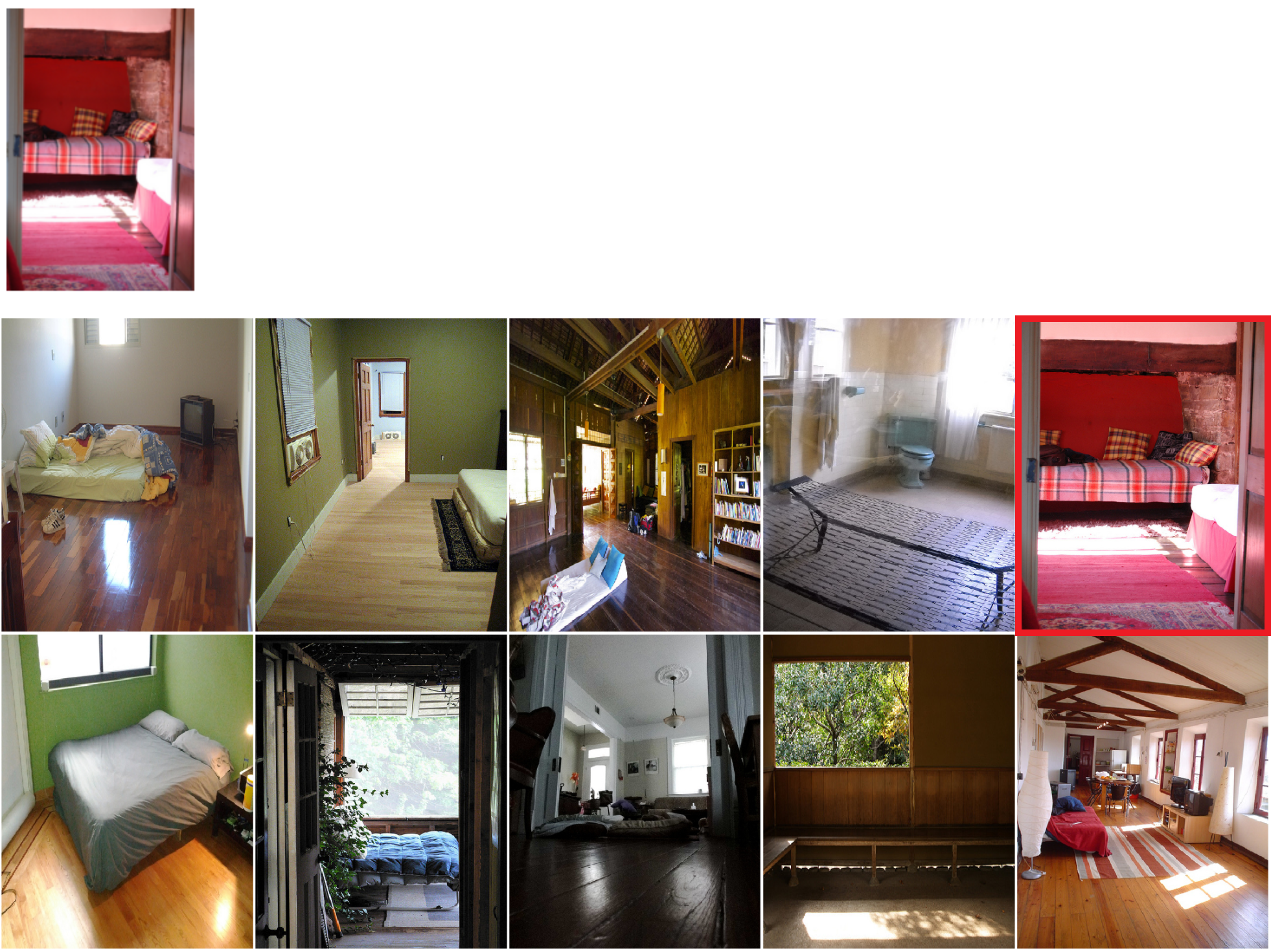}}
\vspace{-6pt}
\caption{Retrieved top 10 images from the query "The sun hits the floor in a rustic bedroom." (Top picture is the ground truth.)}
\label{fig:example_7}
\vspace{-3mm}
\end{figure*}

\end{document}